# Analysis of Object Detection Models for Tiny Object in Satellite Imagery: A Dataset-Centric Approach

Kailas PS[*], Selvakumaran R[†], Palani Murugan[‡], Ramesh Kumar V[§], Malaya Kumar Biswal M[**]

*Grahaa Space, Bangalore, Karnataka, India - 560086*

In recent years, significant advancements have been made in deep learning-based object detection algorithms, revolutionizing basic computer vision tasks, notably in object detection, tracking, and segmentation. This paper delves into the intricate domain of Small-Object-Detection (SOD) within satellite imagery, highlighting the unique challenges stemming from wide imaging ranges, object distribution, and their varying appearances in bird's-eye-view satellite images. Traditional object detection models face difficulties in detecting small objects due to limited contextual information and class imbalances. To address this, our research presents a meticulously curated dataset comprising 3000 images showcasing cars, ships, and airplanes in satellite imagery. Our study aims to provide valuable insights into small object detection in satellite imagery by empirically evaluating state-of-the-art models. Furthermore, we tackle the challenges of satellite video-based object tracking, employing the Byte Track algorithm on the SAT-MTB dataset. Through rigorous experimentation, we aim to offer a comprehensive understanding of the efficacy of state-of-the-art models in Small-Object-Detection for satellite applications. Our findings shed light on the effectiveness of these models and pave the way for future advancements in satellite imagery analysis.

## I. Introduction

In recent years, significant advancements in deep learning algorithms have transformed fundamental computer vision tasks, such as object detection, tracking, and segmentation. Within this landscape, Small Object Detection (SOD) has emerged as a specialized field aimed at identifying small-sized objects, yet it grapples with numerous challenges. These challenges stem from the inherent complexity of learning representations from limited areas and the absence of problem-specific datasets tailored for effective training. Unlike ground-based cameras capturing natural images in a horizontal view, satellite imagery provides a bird's-eye view with a wide imaging range, encompassing comprehensive information. However, for complex landscapes and urban environments, the distribution of foreground items and background information is imbalanced. Small objects, compared to larger ones, often lack sufficient distinguishing features, making them challenging to discern from the background or similar objects. Moreover, the visual attributes of objects in satellite images can vary significantly due to factors such as angle, lighting, and occlusion. Deep neural networks encounter various obstacles in detecting small objects, including inadequate information in individual layers, limited context for small objects, and class imbalances in datasets. Despite the success of many deep learning models in detecting medium to large objects in satellite images, their performance tends to degrade significantly when detecting small objects such as ships and airplanes. To address these challenges, this research aims to contribute to the field of Small Object Detection by conducting a comprehensive review of the latest state-of-the-art deep learning models, specifically tailored for small object detection in satellite imagery. We have curated a dataset comprising 3000 images representing three classes—cars, ships, and airplanes—to train complex models in resource-constrained environments. Furthermore, satellite video-based object tracking is crucial for applications like military surveillance and environmental monitoring. However, it poses unique challenges such as small target sizes, low frame rates, and illumination changes. We evaluate tracking performance using the Byte Track algorithm on segments of videos from the SAT-MTB dataset for the same classes. The results are discussed comprehensively in the final section of the paper, providing insights into the effectiveness of the proposed approaches and avenues for future research.

---

[*] Project Intern, Grahaa Space, Bangalore, Karnataka, India – 560086. Non-Member AIAA
[†] Project Intern, Grahaa Space, Bangalore, Karnataka, India – 560086. Non-Member AIAA.
[‡] Senior Technology Consultant, Grahaa Space, Bangalore, Karnataka, India – 560086. Non-Member AIAA.
[§] Founder & CEO, Grahaa Space, Bangalore, Karnataka, India – 560086. Non-Member AIAA. **Contact: rkvconf@gmail.com**
[**] Senior Research Scientist, Grahaa Space, Bangalore, Karnataka, India – 560086. Non-Member AIAA. **Contact: malay@grahaa.com**.

## II. Literature Review

Computer vision has improved significantly with the combination of deep learning and methods such as Region-based Convolutional Neural Networks (RCNN) and Feature Pyramid Networks (FPN). These methods have enhanced the accuracy and speed of image and video analysis. Researchers have developed state-of-the-art models such as Faster-RCNN and RetinaNet for object detection, thanks to the availability of large-scale image datasets like ImageNet, MS-COCO, and OTB. Backbone networks like VGGNets, ResNets, Inception, and DenseNet, regional proposal methods like RPN and selective search, anchor boxes, and loss functions such as SmoothL1 and Focal Loss are the key components that enhance the accuracy and efficiency of object detection models. Two benchmarks were introduced by Cheng et al. in 2023, specifically developed for aerial object detection, called SODA-A and SODA-D. These benchmarks feature very high-resolution images annotated with OBB. Various innovative methods have been implemented, such as sample-oriented techniques, scale-aware approaches, attention-based models, feature-imitation methods, context-modelling strategies, and focus-and-detect methodologies, each aiming to enhance the accuracy and efficiency of small object detection by addressing specific inherent issues. ConvNext-T emerges as the preferred backbone in different scenarios. SODA-A evaluation highlights RoI Transformer's top performance, particularly due to its effective proposal generator.

Han et al. proposed a novel Unmanned Aerial Vehicle (UAV) object detection dataset (UAVOD-10) and a context-scale-aware detector (CSADet), emphasizing small and weak object detection in UAV images. The dataset encompasses 844 images and 18,234 instances labeled with horizontal bounding boxes, including common targets like buildings and vehicles, as well as novel targets such as cable towers, quarries, and landslides. The existence of large-scale objects can significantly impact the predictions and evaluation metrics of the models. CSADet employs a feature extractor and incorporates a context-aware block that utilizes deformable convolution layers and non-local-based global context blocks. Additionally, a multiscale feature refinement module is introduced to capture high-level semantic and low-level images, addressing the challenges encountered by multiscale objects in high-resolution remote sensing images. Xu et al. introduced the Normalized Gaussian Wasserstein Distance (NWD) measure to assess similarity between small object bounding boxes, addressing the limitations of Intersection Over Union (IoU). Furthermore, the novel approach introduced is validated against the dataset AI-TOD-v2, which is the enhanced annotated version of earlier AI-TOD. Strategies for small object detection can be broadly categorized into multi-scale feature learning, context-based learning, data augmentation, improved training procedures, and label assignment strategies. These methods address challenges related to small objects, such as limited feature information, scale variations, and imbalance between positive and negative samples. This imbalance, particularly for small objects, delays performance. NWD-RKA strategy addresses this issue by increasing positive samples during training, providing better supervision for the network.

Liu et al. discussed key challenges in deep learning for image-based object detection and ways of mitigating these issues. The authors bring valuable insights into detecting objects in aerial images, which involves various techniques such as addressing orientation challenges with Rotation-Invariant CNNs and anchor rotation methods, incorporating context information using feature map concatenation and dilated convolutions, tackling class imbalance through IoU-guided frameworks, and employing multi-scale models like Multi-Scale and Rotation-Insensitive Convolutional Channel Features (MsRi-CCF).

Zhang et al. discuss various methods for tracking traffic objects in satellite videos and categorize tracking methods into four types: Correlation Filters (CF-based), Tracking-By-Detection (TBD), Deep Learning (DL-based), and optical flow-based. The CF-based methods utilize correlation filters and have shown improvements in tracking accuracy. To enhance evaluation, a new multi-level tracking benchmark (MLTB) dataset based on Wright Patterson Air Force Base (WPAFB) is introduced, featuring 401 vehicle tracks categorized by difficulty, aiding the development of remote sensing tracking methods. The authors explore tracking objects in satellite videos, distinguishing between artificial targets (traffic and ships) and natural targets (typhoons, fire, and ice). It highlights the need for different datasets and tracking algorithms based on the characteristics of these targets, emphasizing the challenges of occlusion.

### A. Dataset

The application of transfer learning has yielded remarkable outcomes in computer vision. However, the direct transfer of deep learning-based object detection methods to optical remote sensing images presents challenges. Additionally, existing publicly available datasets often suffer from limitations such as small-scale image numbers and object categories, hindering the advancement of deep learning-based object detection methods.

For instance, the LEVIR dataset, released in 2018, comprises over 22,000 images of 800 x 600 pixels, predominantly covering ground scenes like urban, rural, mountainous, and marine environments, featuring three categories: aircraft, ships, and oil tanks. Similarly, the DOTA dataset, published in 2018, ranges in image sizes from $800 \times 800$ to $20,000 \times 20,000$ pixels and contains 15 common categories. Subsequent versions like DOTA-v1.5 and DOTA-v2.0 introduced additional annotations for extremely small instances and collected more diverse imagery sources.

The SODA-A dataset, introduced in 2020, serves as a benchmark dataset with annotations for various objects like airplanes, helicopters, vehicles, ships, and more. However, its high resolution and large size demand substantial computational resources. The AI-TOD dataset, also released in 2020, offers object detection instances across aerial images, addressing annotation and location errors in its second version (AI-TOD-v2).

Despite the richness of these datasets, their wide-ranging dimensions pose significant computation constraints. Resizing these images risks distorting spatial information, while the presence of large objects in the datasets may skew evaluation metrics like mAP. Thus, there's a need for datasets specifically focusing on tiny objects to evaluate algorithms accurately. This necessity led to the creation of the SkyFusion dataset, which aligns with the MS-COCO's metric for tiny objects, defining them as those occupying areas less than or equal to $32 \times 32$ or 1024 pixels. The average area occupied by objects in the SkyFusion dataset falls within this threshold, enabling precise evaluation of algorithms' performance in detecting tiny objects.

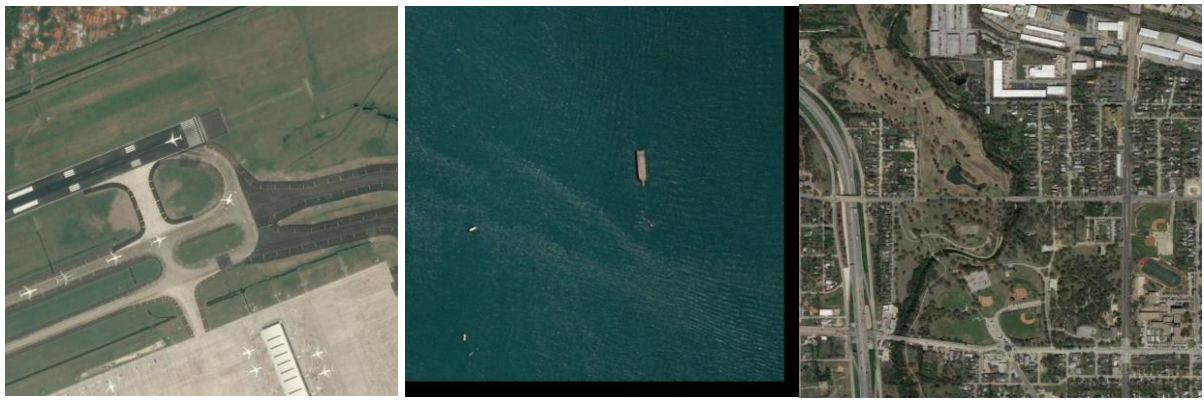

**Figure-1 Satellite Imaging and Object Detection**

### III. Preprocessing and Experimental Setup

The Sky Fusion dataset, a cornerstone of our research, was meticulously curated to focus solely on tiny objects in satellite images. We began by sampling over 2000 images from the vehicle and ship classes of the AiTODv2 dataset, ensuring a balanced distribution of around 1000 images per class. These images were then complemented with data from the Airbus Aircraft Detection dataset, which was initially tiled to generate over 2000 aerial images showcasing aircraft in airports, each with a size of 640x640 pixels. Subsequently, approximately 1000 images were sampled from this generated set to maintain uniformity across classes.

The synthesis of this dataset aimed to provide a precise evaluation platform for algorithms' ability to detect tiny objects in satellite imagery. To facilitate model training, we utilized Google Colab with a Tesla T4 GPU boasting 15GB of RAM, complemented by 12GB of RAM. The algorithms were sourced directly from the model zoo of MM Detection, a popular object detection framework. Following the creation of the subsample, the data underwent compression and annotation standardization to MS-COCO standards using Roboflow, ensuring compatibility and consistency across the dataset.

### IV. Scientific Methodology

**A. Faster R-CNN**

The introduction of region-based Convolutional Neural Network (R-CNN) marked a significant advancement in object detection, ushering in a two-stage framework involving region proposal generation and subsequent classification. However, its reliance on computing numerous regional proposals resulted in computational inefficiency, necessitating the development of more streamlined solutions. Fast R-CNN addressed this inefficiency by performing feature extraction over the entire image using a Region of Interest (RoI) layer, thereby improving processing speed compared to traditional R-CNNs. Despite this improvement, Fast R-CNN still required pre-generated region proposals, typically produced using hand-engineered techniques like EdgeBox and selective search, thereby creating a bottleneck in the object detection process.

To address this limitation, Faster R-CNN was proposed, combining region proposal generation and object recognition to streamline the computational process. The key innovation lies in sharing convolutional layers between the Region Proposal Network (RPN) and the Fast R-CNN object detector, allowing for efficient feature extraction by passing the image through the CNN only once. The framework comprises two modules: the RPN for rapid region proposal generation and the Fast R-CNN for classification and bounding box refinement. By sharing convolutional layers, computational efficiency is significantly enhanced, enabling the utilization of deep CNN models such as VGG16, ResNet, or Inception for feature extraction.

Additionally, Faster R-CNN employs two loss functions—classification loss (cross-entropy) and regression loss (smooth L1)—to further refine object localization accuracy by fine-tuning proposed locations and anchor boxes. This cohesive architecture not only improves computational efficiency but also enhances object detection accuracy, making it a pivotal advancement in the field of computer vision

**B. RetineNet**

RetinaNet is a one-stage object detector that challenges conventional two-stage detectors such as Faster R-CNN. It tackles the issue of class imbalance by training with a novel loss function called Focal Loss, which achieves state-of-the-art performance without the need for complex sampling methods or hard example mining techniques. Focal Loss was introduced to address class imbalance by reshaping the cross-entropy loss to down-weight easy samples and focus on hard negatives, with a modulation factor based on the tuning parameter as denoted in equation 1. The focal loss, denoted as $FL(p_t)$, is an alpha-balanced variant of the cross-entropy loss, defined as:

$$FL(p_t) = -\alpha_t(1 - p_t)^\gamma \log(p_t) \qquad (1)$$

In instances of misclassification where $p_t$ is small, the modulating factor remains close to 1, leaving the loss unaffected. As $p_t$ approaches 1, the factor tends to 0, downweighting the loss for well-classified examples. The focusing parameter, marked by γ, easily adjusts the rate at which easy examples are downweighted. Compared to balanced cross-entropy, it is effective in handling imbalanced datasets, reducing the need for traditional two-stage cascade mechanisms in detectors.

The RetinaNet model integrates a Feature Pyramid Network (FPN) above its backbone and two subnetworks for object classification and bounding box regression. The FPN constructs a multi-scale feature pyramid that employs anchor boxes with specific aspect ratios and scales. The anchor assignments and box regression targets are determined using the intersection-over-union (IoU) threshold. A small Fully Connected Network (FCN) is used to predict the classification probability of object labels with shared parameters across pyramid levels. The design encompasses convolutional layers with ReLU activations and a final layer with sigmoid activations. The box regression subnet, which is attached to each pyramid level, utilizes a similar FCN structure to predict anchor box offsets. A crucial design consideration is anchoring usage, addressing the density of coverage across possible image boxes. RetinaNet-101-600 matches the accuracy of ResNet-101-FPN Faster R-CNN but runs 50% faster.

**C. FCOS**

Object detection traditionally involves recognition and localization, often relying on anchor-based methods. However, these methods pose challenges in terms of hyperparameter tuning, handling size variations, and computational complexity. FCOS [45] introduced an anchor-free approach using Fully Convolutional Networks to predict adjacent pixels, integrating object detection with other vision tasks. It outperforms anchor-based detectors such as RetinaNet, YOLO [46], and SSD [47], while having lower design complexity by avoiding anchor boxes. FCOS simplifies training and avoids the complex computations involved in anchor boxes, showcasing state-of-the-art performance and potential for broader instance-level recognition tasks.

Unlike anchor-based detectors, FCOS directly regresses bounding boxes at each location, treating locations as training samples. Positive samples are determined if a location falls within the designated ground-truth boxes, and the regression targets are computed appropriately as denoted in equation 2 below.

$$L(\{\boldsymbol{p}_{x,y}\}, \{\boldsymbol{t}_{x,y}\}) = \frac{1}{N_{\text{pos}}}\sum_{x,y} L_{\text{cls}}(\boldsymbol{p}_{x,y}, c^*_{x,y})$$
$$+ \frac{\lambda}{N_{\text{pos}}}\sum_{x,y} \mathbb{1}_{\{c^*_{x,y}>0\}} L_{\text{reg}}(\boldsymbol{t}_{x,y}, \boldsymbol{t}^*_{x,y}) \quad (2)$$

The loss function L combines focal loss for classification ($L_{\text{cls}}$) and IOU loss for regression ($L_{\text{reg}}$) over positive samples, with a balancing weight λ. The indicator function ensures that the summation is performed only for locations with positive samples, and N_pos represents the number of positive samples in the feature maps. The indicator function $\mathbb{1}_{\{c^*_{x,y}>0\}}$ returns 1 if $c^*_i > 0$ and 0 otherwise. It utilizes multi-level prediction to enhance recall and address ambiguity from overlapping bounding boxes. Additionally, a "centerness" branch is introduced to significantly reduce low-quality detections, thereby improving overall performance. The detector identifies locations as training samples rather than anchor boxes, simplifying the process and achieving better results than anchor-based counterparts. The network outputs consist of classification labels and bounding box coordinates, whereas anchor-based detectors are less dependent on variables.

**D. Varifocal Net**

Verifocal Net fundamentally combines the FCOS detector with the ATSS [49] training strategy. This model identifies limitations in using the predicted centerness score or IoU with classification score and addresses them by replacing the classification score with an IoU-aware classification score (IACS). This enables Verifocal Net to better judge detections, highlighting the importance of accurately selecting high-quality detections.

The model introduces Varifocal Loss, inspired by Focal Loss, for training dense object detectors to predict an IoU-aware Classification Score (IACS). Unlike Focal Loss, Varifocal Loss treats positive and negative examples asymmetrically, focusing on hard negatives and high-quality positives, thus addressing class imbalance during training. The loss is dynamically scaled based on the predicted IACS and the target score, emphasizing learning signals from rare positive examples with high IoU.

VFNet incorporates three innovative components: varifocal loss, star-shaped bounding box feature representation, and bounding box refinement. Varifocal loss is also based on the binary cross-entropy loss, as defined in equation 3 below.:

$$\text{VFL}(p, q) = \begin{cases} -q(q\log(p) + (1-q)\log(1-p)) & q > 0 \\ -\alpha p^{\gamma} \log(1-p) & q = 0 \end{cases} \quad (3)$$

Where $p$ represents the predicted IACS, and $q$ is the target score. "For a foreground point, the value of q for its ground-truth class is determined by the IoU between the generated bounding box and its ground truth (gt IoU). Meanwhile, for background points, the target q for all classes is uniformly set to 0. The training of VFNet is supervised by the loss function denoted in equation 4:

$$\text{Loss} = \frac{1}{N_{\text{pos}}}\sum_i \sum_c \text{VFL}(p_{c,i}, q_{c,i})$$
$$+ \frac{\lambda_0}{N_{\text{pos}}}\sum_i q_{c^*,i} L_{\text{bbox}}(\text{bbox}'_i, \text{bbox}^*_i) \quad (4)$$
$$+ \frac{\lambda_1}{N_{\text{pos}}}\sum_i q_{c^*,i} L_{\text{bbox}}(\text{bbox}_i, \text{bbox}^*_i)$$

The overall loss function is defined as a combination of three terms: the Varifocal Loss (VFL) applied to predicted and target IACS ($p_{c,I}$ and $q_{c,i}$), the GIoU loss[36] ($L_{\text{bbox}}$) weighted by training targets ($q_{c^*,i}$) for initial and refined bounding boxes, and the ATSS-based foreground and background point identification. The balance weights ($\lambda_0$ and $\lambda_1$) for the GIoU loss are empirically set to 1.5 and 2.0, respectively. The normalization factor $N_{\text{pos}}$ is determined by the number of foreground points, ensuring the total loss is appropriately scaled. These elements collectively contribute to advancing detection capabilities and refining the overall detection process.

### E. FoveaBox

Following the philosophy of Varifocal Net and FCOS, FoveaBox operates on the principle of an anchor-free detector, achieving state-of-the-art results without predefined candidate boxes. Unlike anchor-based systems, FoveaBox can predict object locations and categories, simplifying training and allowing flexibility.

Mathematically each ground-truth bounding box in FoveaBox is specified as $G = (x_1, y_1, x_2, y_2)$. Starting from a positive point $(x, y)$ in $R^{pos}$, FoveaBox directly computes the normalized offset between $(x, y)$ and four boundaries as in the equation 5:

$$\begin{aligned} t_{x_1} &= \log \frac{s_l(x+0.5)-x_1}{r_l}, \\ t_{y_1} &= \log \frac{s_l(y+0.5)-y_1}{r_l}, \\ t_{x_2} &= \log \frac{x_2-s_l(x+0.5)}{r_l}, \\ t_{y_2} &= \log \frac{y_2-s_l(y+0.5)}{r_l}. \end{aligned} \quad (5)$$

The above function initially transforms the coordinate (x, y) to the input image and further calculates the normalized offset between the projected coordinate G. Subsequently, it normalizes the targets using a log-space function, where $r_l$ denotes the fundamental scale. FoveaBox utilizes anchor-based techniques to define positive/negative samples based on ground-truth boxes, enhancing simplicity.

FoveaBox surpasses state-of-the-art object detection methods, excelling in both one-stage and two-stage detectors. The model, based on anchor-free detection, shows robustness to varying anchor densities, excels in predicting bounding boxes across different aspect ratios, and generates high-quality regional proposals, outperforming traditional RPN methods.

### F. RTMDet

Continuing the efforts to capture the global context and better represent images is the new family of Real-Time Models for Object Detection, named RTMDet [51]. RTMDet is also capable of performing instance segmentation and rotated object detection. It uses large-kernel depth-wise convolutions in the basic building block of the backbone and neck in the model, which improves the model's capability of capturing the global context. The large effective receptive field in the backbone is beneficial for dense prediction tasks like object detection and segmentation as it helps to comprehensively capture and model the image context.

Hence, the model introduces 5×5 depth-wise convolutions in the basic building block of CSPDarkNet [52] to increase the effective receptive fields with Batch Normalization (BN) [53]. The new label assignment strategy, otherwise known as soft label assignment strategy, is based on SimOTA, and its cost function can be formulated as in equation 6.

$$C = \lambda_1 C_{cls} + \lambda_2 C_{reg} + \lambda_3 C_{\text{center}} \quad (6)$$

where $C_{cls}, C_{center}$, and $C_{reg}$ correspond to the classification cost, region prior cost, and regression cost, respectively, and $\lambda_1 = 1, \lambda_2 = 3$, and $\lambda_3 = 1$ are the weights of these three costs by default. RtmDet also utilizes the logarithm of the IoU as the regression cost instead of GIoU used in the loss function, amplifying the cost for matches with lower IoU values. Additionally, RtmDet follows YOLOX [54] in their training strategy, excluding the data augmentation step. It introduces the use of Large-Scale Jittering (LSJ) [55], allowing for fine-tuning of the model in a domain more closely aligned with real data distributions.

### G. DDOD

The model DDOD [56] employs a training pipeline similar to ATSS for enhancing the training of dense object detectors. The model addresses three key issues in training implementation: label assignment, localization, and class imbalance.

Historically, the regression loss has been applied only to "foreground" samples in classification. One widely accepted practice in label assignment is to regress samples that are assigned as positives in the classification branch. The DDOD model adopts an innovative label assignment methodology, which can be mathematically represented as follows:

Given an image I and suppose there are P predictions based on the predefined anchors and N ground truths. For each candidate (anchor or center point) $P_i$, the foreground probability $\hat{p}(i)$ and the regressed bounding box $\hat{b}(i)$ are output with respect to each category. To this end, a Cost Matrix can be formulated as shown in equation 7.

$$C_{i,\pi(i)} = \underbrace{\mathbb{1}[\pi(i) \in \Omega_i]}_{\text{spatial prior}} \cdot \underbrace{\left(\widehat{p_{\pi(i)}}(i)\right)^{1-\alpha}}_{\text{classification}} \cdot \underbrace{\left(IoU\left(b_i, \widehat{b_{\pi(i)}}(i)\right)\right)^{\alpha}}_{\text{regression}} \qquad (7)$$

where $C_{i,\pi(i)} \in [0,1]$ represents the matching quality of the $\pi(i)$-th prediction with respect to each ground truth, and $\Omega_i$ denotes the set of candidate predictions for i-th ground truth. Although classification and regression (localization) share the same receptive fields, they have different preferences: classification desires regions with rich semantic information, while regression prefers to focus on edge parts [57]. Consequently, optimal performance cannot be guaranteed with the same receptive fields.

The pyramid layers exacerbate the class imbalance problem due to the equal supervision on all training samples. To address this issue, DDOD utilizes the FPN hierarchical loss, which assigns more weight to samples in deeper layers during training. This approach aims to improve object training across all pyramid levels by mitigating the gradient imbalance caused by varying sample counts in different layers

**H. DetectoRS**

Modern object detectors achieve remarkable performance through a "look and think twice" strategy. DetectoRS explores this approach in designing the backbone for object detection. It introduces Recursive Feature Pyramid (RFP), expanding upon the conventional Feature Pyramid Networks (FPN). RFP incorporates feedback connections into FPN, enhancing the bottom-up backbone layers with additional feedback from FPN layers. This iterative process strengthens FPN, gradually enhancing the representations to be more powerful.

Let $B_i$ denote the i-th stage of the bottom-up backbone, and $F_i$ denote the i-th top-down FPN operation. The resulting feature maps by backbone equipped with FPN outputs $\{f_i \mid i = 1, ..., S\}$ where $S$ is the number of the stages. the output feature $f_i$ is defined by.

$$f_i = F_i(f_{i+1}, x_i), \quad x_i = B_i(x_{i-1})$$

where $x_0$ is the input image and $f_{S+1} = 0$. The object detector built on FPN uses $f_i$ for the detection computations.

Recursive Feature Pyramid (RFP) adds feedback connections to FPN. Let $R_i$ denote the feature transformations before connecting them back to the bottom-up backbone. Then, $\forall i = 1, ..., S$, the output feature $f_i$ of RFP is defined by.

$$f_i = F_i(f_{i+1}, x_i), \quad x_i = B_i\left(x_{i-1}, R_i(f_i)\right)$$

which makes RFP a recursive operation.

In convolutional layers, atrous convolution effectively extends the filter's field of vision. The SAC architecture includes two small global context modules, positioned before and after the SAC component, to compress input features using global average pooling.

Moreover, the model features a unique Switchable Atrous Convolution, which utilizes spatially dependent switch functions to consolidate results from convolving the same input feature with different atrous rates. These switch functions regulate SAC outputs, varying across feature map sites. SAC enhances the detector by transforming each typical 3x3 convolutional layer in the bottom-up backbone, significantly improving performance. These mechanisms are integral to the HTC model, significantly enhancing performance while maintaining speed.

## V. Results and Discussion

**Table-1 Comparison of data from various Machine Learning Models**

| Model | Variant | mAP | mAP$_{50}$ | mAP$_{75}$ | mAP$_S$ | mAP$_M$ | mAP$_L$ |
|---|---|---|---|---|---|---|---|
| **DetectoRS** | ResNet50-RFP | 0.123 | 0.275 | 0.091 | 0.114 | 0.321 | -1.000 |
| | ResNet50 -SAC | 0.134 | 0.298 | 0.102 | 0.125 | 0.343 | -1.000 |
| **DDOD** | ResNet50 + FPN | **0.339** | **0.611** | **0.341** | **0.300** | 0.436 | -1.000 |
| | ResNet101 + FPN | 0.337 | 0.601 | 0.337 | 0.297 | 0.456 | -1.000 |
| **Faster RCNN** | ResNet50 + FPN | 0.103 | 0.235 | 0.076 | 0.094 | 0.315 | -1.000 |
| | ResNet101 + FPN | 0.103 | 0.240 | 0.074 | 0.093 | 0.322 | -1.000 |
| **FCOS** | ResNet50 + FPN | 0.300 | 0.547 | 0.332 | 0.263 | 0.460 | -1.000 |
| **FoveaBox** | ResNet50 + FPN | 0.290 | 0.512 | 0.309 | 0.245 | 0.453 | -1.000 |
| | ResNet101 + FPN | 0.283 | 0.491 | 0.302 | 0.238 | 0.426 | -1.000 |
| **RetinaNet** | ResNet50 + FPN | 0.296 | 0.524 | 0.306 | 0.248 | 0.450 | -1.000 |
| | ResNet101 + FPN | 0.292 | 0.527 | 0.297 | 0.244 | 0.465 | -1.000 |
| **RTMDet** | RTMDet-Tiny | 0.248 | 0.424 | 0.274 | 0.205 | 0.417 | -1.000 |
| | RTMDet-Small | 0.259 | 0.448 | 0.283 | 0.217 | 0.432 | -1.000 |
| | RTMDet-Medium | 0.272 | 0.471 | 0.288 | 0.233 | 0.434 | -1.000 |
| | RTMDet-L | 0.286 | 0.501 | 0.302 | 0.250 | 0.438 | -1.000 |
| **Varifocal Net** | ResNet50 + FPN | 0.182 | 0.455 | 0.113 | 0.173 | 0.358 | -1.000 |
| | ResNet101 + FPN | 0.179 | 0.438 | 0.105 | 0.173 | 0.330 | -1.000 |

In this study, we meticulously curated a dataset comprising 3000 satellite images showcasing cars, ships, and airplanes, aiming to evaluate state-of-the-art deep learning models for Small-Object-Detection (SOD) within satellite imagery. Utilizing Google Colab with a Tesla T4 GPU, we rigorously trained and evaluated models sourced from the MMDetection model zoo, including Faster R-CNN, RetinaNet, FCOS, Varifocal Net, FoveaBox, RTMDet, DDOD, and DetectoRS. Our investigation revealed significant advancements in addressing the challenges of small object detection, with models like RetinaNet leveraging Focal Loss to tackle class imbalance and anchor-free approaches such as FCOS and FoveaBox simplifying training while achieving robustness to varying anchor densities. Innovative approaches like Varifocal Net and RTMDet showcased enhanced object detection capabilities through advanced loss functions and feature representations. These findings underscore the importance of tailored approaches for small object detection in satellite imagery and pave the way for future advancements in this critical domain..

### A. Byte Track

Given these models, the image likely compares the mAP performance of these different object detection models on a specific dataset as the number of training epochs increases (x-axis). The goal would be to see which model achieves the best mAP (highest value on the y-axis) and how quickly it improves during training.

By comparing the curves for each model, we can see:

1) Overall mAP performance: The highest curve on the graph at the end of training (furthest right on the x-axis) represents the model with the best overall mAP.

2) Training speed: The rate at which each curve increases can indicate how quickly each model improves with training. A steeper curve suggests faster improvement.

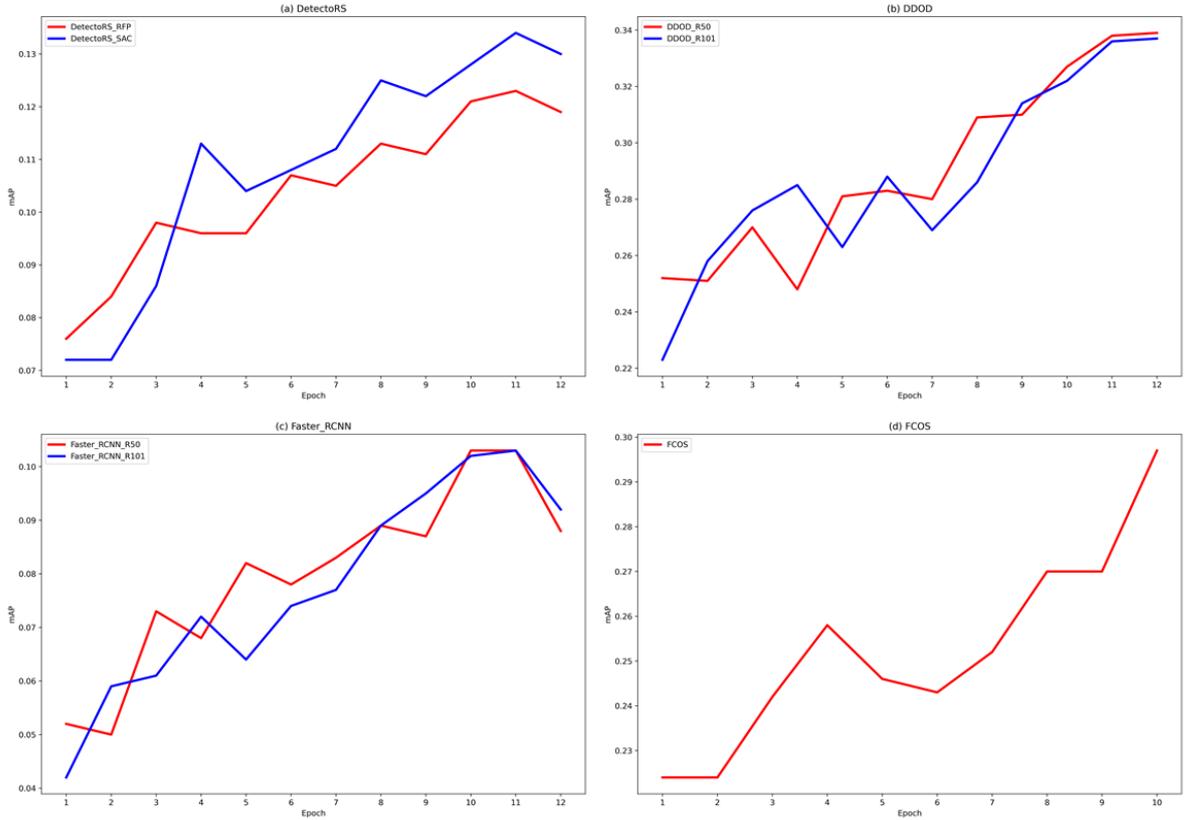

**Figure-2 Comparison of Epoch and mAP (Mean Average Position) using various data models.**

## VI. Conclusion and Future Works

Inspired by the challenge of detecting large and small objects at different scales, Singh and Davis introduced a training strategy called Scale Normalization for Image Pyramids (SNIP) [25]. SNIP selectively back-propagates gradients of object instances of various sizes by utilizing all ground truth boxes to assign labels to proposals during training. They train the classifier by selecting only ground truth boxes and proposals falling within a specified size range at a particular resolution. Similarly, all ground truth boxes are used to assign labels to anchors for training the Region Proposal Network (RPN).

Later, Singh et al. proposed SNIPER [26], another approach for efficient multi-scale training. SNIPER processes context regions around ground truth instances at the appropriate scale instead of the whole image pyramid, accelerating multi-scale training by sampling low-resolution regions. Small object detection remains a challenging task in computer vision, and this paper provides a comprehensive review of small object detection methods based on deep learning. However, there is still much work to be done, with a focus on the following aspects.

## VII. Conflict of Interest

The authors declare no conflict of interest.

## VIII. Funding

No external funding was received to support or conduct this study